\documentclass[conference]{IEEEtran}
\usepackage{times}
\usepackage{epsfig}
\usepackage{graphicx}
\usepackage{amsmath}
\usepackage{amssymb}
\usepackage{verbatim}
\usepackage{float}
\usepackage{multirow}

\ifCLASSINFOpdf
\else
\fi

\begin{document}
\title{Implicit Language Model in LSTM for OCR}

\author{\IEEEauthorblockN{Ekraam Sabir}
\IEEEauthorblockA{Information Sciences Institute\\
University of Southern California\\
Marina Del Rey, California 90007\\
Email: esabir@isi.edu}
\and
\IEEEauthorblockN{Stephen Rawls}
\IEEEauthorblockA{Information Sciences Institute\\
University of Southern California\\
Marina Del Rey, California 90007\\
Email:srawls@isi.edu}
\and
\IEEEauthorblockN{Prem Natarajan}
\IEEEauthorblockA{Information Sciences Institute\\
University of Southern California\\
Marina Del Rey, California 90007\\
Email: pnataraj@isi.edu}}


%


\maketitle

\begin{abstract}
Neural networks have become the technique of choice for OCR, but many aspects of how and why they deliver superior performance are still unknown. One key difference between current neural network techniques using LSTMs and the previous state-of-the-art HMM systems is that HMM systems have a strong independence assumption. In comparison LSTMs have no explicit constraints on the amount of context that can be considered during decoding. In this paper we show that they learn an implicit LM and attempt to characterize the strength of the LM in terms of equivalent n-gram context. We show that this implicitly learned language model provides a 2.4\% CER improvement on our synthetic test set when compared against a test set of random characters (i.e. not naturally occurring sequences), and that the LSTM learns to use up to 5 characters of context (which is roughly 88 frames in our configuration). We believe that this is the first ever attempt at characterizing the strength of the implicit LM in LSTM based OCR systems.
\end{abstract}


%
\IEEEpeerreviewmaketitle

\section{Introduction}
At the heart of any Optical Character Recognition (OCR) system is a glyph recognition model whose purpose is to identify individual glyphs based on extracted features. However, in addition to the glyph model, OCR systems typically employ feature extraction, segmentation and language modeling modules to get competitive performance \cite{plamondon_online_2000}. Of these, language modeling improves the output of a glyph recognition model conditioned on the distribution of characters or words from task-relevant-text i.e. a language model. A language model can build a word or a character language model \cite{koerich_large_2003-1}. \cite{kukich_techniques_1992} gives a general survey on the use of language modeling. Irrespective of the OCR method, language modeling has been investigated independently and has also played a crucial role in achieving better performance \cite{farooq_phrase_2008}\cite{bunke_offline_2004}.

Hidden Markov Model (HMM) based systems provided segmentation free OCR and outperformed then existing segmentation based approaches \cite{margner_arabic_2007}\cite{natarajan_multi-lingual_2008}. This conditional independence limitation was addressed by Recurrent Neural Networks (RNNs) \cite{elman_finding_1990} which theoretically have no limitations on the length of context they can utilize.

While Neural Networks have been used with success for OCR in the past \cite{lecun_backpropagation_1989}\cite{lecun_gradient-based_1998}, it is only recently that recurrent networks and particularly LSTMs \cite{hochreiter_long_1997} became popular for the OCR task, improving upon the performance of HMM OCR systems \cite{breuel_high-performance_2013}\cite{graves_offline_2009}. Solutions to the vanishing and exploding gradient problems associated with the training of RNNs \cite{pascanu_difficulty_2013}\cite{hochreiter_long_1997} coupled with the introduction of Connectionist Temporal Classification (CTC) loss \cite{graves_connectionist_2006} played a major role in this resurgence. The CTC loss was particularly well suited to tackling the OCR problem, removing the necessity for frame level label assignment.

The convincing performance improvements made by LSTMs however stand in stark contrast to the limited interpretability of these networks. The functionality of individual neurons, weights and to some extent the hidden layers themselves remains ambiguous at best. To this end, a significant amount of effort has been expended in explaining LSTMs regarding their structure such as performance of LSTM variants with and without hyperparameter tuning and the effects of depth \cite{greff_lstm:_2016}\cite{zaremba_empirical_2015}\cite{pascanu_how_2013}. \cite{hermans_training_2013} and \cite{karpathy_visualizing_2015} explore the memory and functionality of neurons with experiments that test long term reasoning among others, on a character language model learning task.

Continuing in the general direction of unraveling LSTMs, we explore their possibility of learning a language model when trained on a different but related OCR task. Foundational credibility for LSTMs learning an internal language model when trained for OCR can be enumerated from previous discussion as follows: 1) LSTMs do not have an explicit restriction on the amount of context they can learn; 2) They have been shown to learn character language models when trained for it specifically as in \cite{karpathy_visualizing_2015}; and 3) Learning a language model in general helps improve performance on the OCR task. We find additional evidence for this idea in \cite{ul-hasan_can_2013} where an LSTM is trained on a multilingual OCR task. The setup involves testing multiple LSTM models which are trained on one native language and tested on other foreign languages with the same glyphs. The results on a real world problem show up to 3.6\% CER difference in performance when testing on foreign languages, which is indicative of the model's reliance on the native language model. However, the authors of \cite{ul-hasan_can_2013} do not explain this phenomena.

In this paper we attempt to advance our scientific understanding of LSTMs, particularly the interactions between language model and glyph model present within an LSTM. We call this internal language model the implicit language model (implicit LM). Our contributions in this paper include: 1) Establishing the presence of implicit LM under controlled conditions; and 2) characterizing the nature of implicit LM by finding how many characters of context it makes use of. The implicit LM we characterize is different from the language model in \cite{hermans_training_2013}\cite{karpathy_visualizing_2015} discussed above in that the setting and requirement for learning a language model is different: OCR explicitly requires learning a glyph model instead of a language model. A recent benchmarking paper on the use of LSTM for OCR \cite{breuel_benchmarking_2015} has not covered this and to the best of our knowledge has also not been covered in literature.


\section{METHODOLOGY}

\begin{figure}[t]
\centering
\includegraphics[width=0.9\linewidth,keepaspectratio]{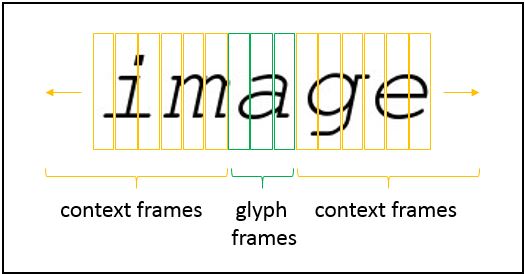}
\caption{A visualization of the context and glyph frames for identifying the glyph \textit{a} with respect to surrounding characters. Each frame is an OCR input for one time-step.}
\end{figure}

The implicit LM is a learned aspect of the LSTM, whose contextual extent cannot be evaluated by known methods or metrics. It is also intertwined with the LSTM's learned glyph model, which is expected to rely on the glyph frames of a character, while the implicit LM relies on the context frames as shown in Figure 1. Any measure of performance on the OCR task is a result of both aspects of the model working in unison to predict a character. In such conditions, it is not possible to isolate the contributions of implicit LM by a single measure of performance on any data. This makes analysis of implicit LM a challenging task and any approach to distinguish its contributions has to be novel. We attempt to characterize the language model by a series of experiments which together give a clearer picture and address the entangled issues described.

An LSTM benefiting from context is nothing new in itself. In OCR, this temporal aspect of an LSTM allows it to take slices of image across variable width characters and recognize it. However, we create synthetic datasets where characters are not connected and a sufficiently complex model could learn to rely on glyph frames alone for recognition. In such conditions we measure performance on test sets which deliberately deviate from the training character language model, to show that an LSTM benefits from assimilated memory of context frames, in addition to glyph frames. This idea is the basis for the Shuffled Character and N-gram experiments in Section 4.

Even though LSTMs do not have an explicit restriction on memory, as with n-gram language models the amount of useful context is limited. When testing on increasing length sequences seen in training, we expect the performance improvement to saturate beyond a certain length. The length at which the improvement plateaus should be the length of character context implicit LM considers in predicting a character. This gives us motivation for the N-gram experiments in Section 4.

Since LSTMs by design learn context over time-steps instead of characters, an argument can be made for measuring implicit LM over time-steps. A time-step based approach however, may not be a consistent indicator across proportional fonts and different font sizes. Additionally, the CTC loss makes character predictions at varying intermediate frames of characters, giving a special blank prediction for all other frames. A control over font size, font type and character to get a reliable estimate of the memory in implicit LM is an extremely constrained setting to perform experiments in. A character level estimate on other hand though not perfect, has the added advantage of being interpretable.

\section{EXPERIMENTAL SETUP}

\subsection{Data}

The experiments we perform require controlled datasets of fixed length sequences with specific requirements, which is easily created from synthetic images but intractable to find in real world data. Fortunately our objective is to advance the understanding of LSTMs and not directly effect an improvement in performance, thus real world data is not explicitly required. Using synthetic images over handwritten image datasets has the added advantage of eliminating image background noise from interfering with experiment results. 

For training, we intend to train a robust model that generalizes reasonably well to variations in fonts, font sizes. We render 32,180 unique sentences from the books \textit{Tale of Two Cities} and \textit{Ivanhoe} with font sizes in the range 8 to 16 and 6 fonts: Californian FB Italic, Garamond, Georgia, Arial, Comic, Courier Italic for each sentence. If the model is found to learn a character language model, it would be from this set of 32,180 sentences. The validation images are from \textit{The Adventures of Tom Sawyer} and have 1585 unique sentences. They are rendered in a similar fashion and with the same fonts as in training images.

We choose test fonts different from training fonts and with large enough error to be reliably measurable. The training fonts when used for testing give near 0\% error. The test fonts are Comic Bold, Times and Arial Narrow. Sample images of all the fonts are shown in Figure 2. The test dataset for shuffled characters experiment in section 4 contains 3742 sentences from \textit{Wuthering Heights}.

For the remaining experiments, multiple test datasets of small case character sequences or N-grams are created. The N-gram may be \textit{Seen} or \textit{Unseen} in training. We consider an N-gram to be \textit{Seen} if it has a frequency above 10 in the unique training sentences and \textit{Unseen} for 0 frequency. N-grams with frequency between 0-10 are ignored. We create additional sets of N-grams which we call \textit{Purely Unseen}. They are N-grams whose sub N-grams are also \textit{Unseen}. For instance an \textit{Unseen} 5-gram \textit{ameoy} has one \textit{Seen} 3-gram \textit{ame} and four \textit{Seen} 2-grams \textit{am}, \textit{me}, \textit{eo} and \textit{oy}. The sequence \textit{bcgpq} is an example of a \textit{Purely Unseen} N-gram. All N-gram test data comprises of 26 small case English alphabets. Even though spaces appear abundantly in training, we remove them since the model might trip on inconsequential sequences of consecutive spaces leading to an exaggerated estimate of implicit LM. Each N-gram test set of some fixed length and \textit{Seen}, \textit{Unseen} or \textit{Purely Unseen} type has up to 10,000 samples, which should be large enough to measure error reliably.

\begin{figure}[t]
\centering
\begin{tabular}{c}
\includegraphics[width=0.6\linewidth]{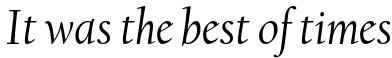}\\
\includegraphics[width=0.6\linewidth]{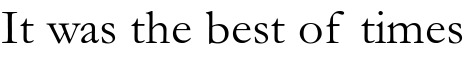}\\
\includegraphics[width=0.6\linewidth]{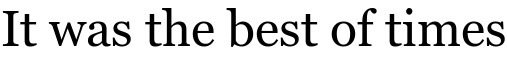}\\
\includegraphics[width=0.6\linewidth]{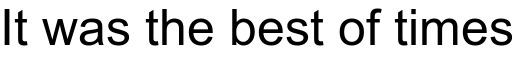}\\
\includegraphics[width=0.6\linewidth]{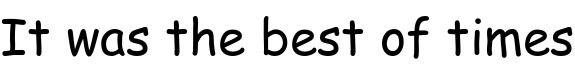}\\
\includegraphics[width=0.6\linewidth]{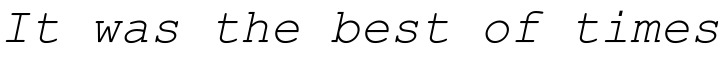}\\
\includegraphics[width=0.6\linewidth]{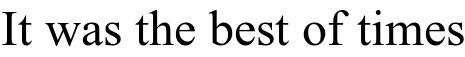}\\
\includegraphics[width=0.6\linewidth]{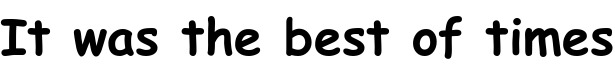}\\
\includegraphics[width=0.6\linewidth]{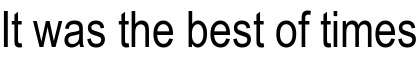}\\
\end{tabular}
\caption{Sample images of the fonts used in training and test. From top to bottom: training: Californian FB Italic, Garamond, Georgia, Arial, Comic, Courier Italic and test: Times, Comic Bold and Arial Narrow}
\end{figure}

\subsection{Preprocessing}

To ensure a constant input size to the model, images are scaled to a constant height of 30 pixels while conserving the aspect ratio. They are also normalized to have zero mean and unit standard deviation as has been recommended in \cite{bishop_neural_1995}.

\subsection{Model}

For a reasonably well performing OCR model, we choose one that is very similar in structure to \cite{rawls_combining_2017} but not fine-tuned like it. The model is segmentation free and the LSTM output does not require processing except for decoding. It takes sliding window image frames of 2 pixel width transformed into a 60x1 vector as raw input features for 2x fully connected layers with 60 units each. All activation functions are Rectified Linear Units (ReLU). It is followed by 2 bidirectional LSTM layers with 256 units in each layer and 700 time-steps, where each time-step is a potential character prediction. The objective function is the CTC loss function \cite{graves_connectionist_2006}. The model architecture is shown in Figure 2.

\begin{figure}[t]
\centering
\includegraphics[width=1\textwidth,height=0.4\textheight,keepaspectratio]{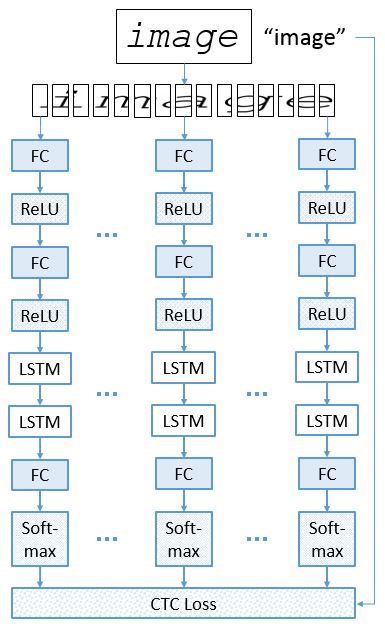}
\caption{An overview of the entire system. The feature extraction layer takes image frame input and the LSTM layers function as recognition layers.}
\end{figure}

\subsection{Training}

The model is trained with a starting learning rate of 0.001 and dropout 0.5. These numbers turn out to be sufficient to train the model and are not fine-tuned. It is trained for over 1 epoch with 0.04\% CER and 0.02\% WER on training and 0.02\% CER and 0.01\% WER on validation.

\subsection{Testing}

We measure error in terms of Character Error Rate (CER) throughout, ignoring Word Error Rate (WER). For comparable CER, the WER of a longer sequence will be inevitably larger than that for a shorter sequence. A single CER is reported on a test dataset of some fixed length containing N-grams of \textit{Seen}, \textit{Unseen} or \textit{Purely Unseen} type.

\section{EXPERIMENTS}

We present the results of experiments in the following subsections. While each experiment on its own is not sufficient to characterize the implicit LM, together they present a more coherent picture.

\subsection{Shuffled Characters Experiment}

In \cite{ul-hasan_can_2013} we see that an LSTMs performance improves by up to 3.6\% CER when using a mixed language model training setup instead of the original character language model. However, the authors of \cite{ul-hasan_can_2013} do not investigate the possibility of an internal language model. We first establish the presence of an implicit LM with an experiment on a controlled dataset. The test dataset for this experiment consists of full length English sentences sampled from \textit{Wuthering Heights} and rendered in test fonts. We shuffle the characters randomly in these sentences and re-render them, resulting in a dataset with same characters in corresponding sentences, but with a random character language model. Ideally the performance between these two sets should be the same, and any difference should come from the implicit LM. Table 1 shows the results from these experiments. Font Arial Narrow shows the biggest difference with CER up to 2.4\% better on normal English language sentences.

\begin{table}[t]
\caption{Difference in cer and wer measure of normal test sentences and shuffled test sentences}
\begin{center}
\begin{tabular}{|c|c|c|c|c|}
	\hline
    \multirow{2}{*}{Font} & \multicolumn{2}{c|}{CER} & \multicolumn{2}{c|}{WER} \\
    \cline{2-5}
    & Normal & Shuffled & Normal & Shuffled \\
    \hline
    Comic bold & 2.4\% & 2.6\% & 6.4\% & 8.9\% \\
    \hline
    Times & 7.2\% & 8.6\% & 25.1\% & 26.5\% \\
    \hline
    Arial Narrow & 0.5\% & 2.9\% & 2.1\% & 6.8\% \\
    \hline
\end{tabular}
\end{center}
\end{table}

\subsection{N-gram Experiment}

We have already established the presence of implicit LM in the shuffled characters experiment. The objective of this experiment is to quantify its contextual limit in terms of characters. As with other language models, the contextual benefits from implicit LM should be limited. Test sets with longer sequences should benefit more from the extra characters up to a certain length after which the implicit LM should saturate in performance. Our hypothesis is that the performance will improve as length increases and plateau where implicit LM stops considering more context frames. We run these experiments on \textit{Seen} 2-gram to 7-gram test sets which derive their language model from the training set. In Table 2 we observe that the performance stops improving beyond 5 characters, indicating that the implicit LM can benefit from up to 5 characters in context for a bidirectional LSTM model. This corresponds to 88 frames of input in our configuration for font size 16 on the widest test font comic bold.

\begin{table*}[t]
\caption{Results on N-gram experiments of Seen, Purely Unseen and Unseen type. The length ranges from 2 to 7}
\centering
\begin{tabular}{|c|c|c|c|c|c|c|}
	\hline
    \multirow{2}{*}{N-gram} &  \multicolumn{3}{c|}{Arial Narrow} & \multicolumn{3}{c|}{Comic Bold} \\
    \cline{2-7}
    & Seen & Unseen & Purely Unseen & Seen & Unseen & Purely Unseen \\
    \hline
    2 & 0.9\% & 4.9\% & 4.9\% & 2.4\% & 6.9\% & 6.9\% \\
    \hline
    3 & 0.5\% & 2.5\% & 5.0\% & 2.3\% & 4.1\% & 7.9\% \\
    \hline
    4 & 0.5\% & 1.9\% & 5.0\% & 2.2\% & 3.6\% & 9.1\% \\
    \hline
    5 & 0.4\% & 1.8\% & 5.0\% & 2.0\% & 3.7\% & 9.4\% \\
    \hline
    6 & 0.4\% & 1.5\% & - & 2.1\% & 3.7\% & - \\
    \hline
    7 & 0.4\% & 1.5\% & - & 2.0\% & 3.7\% & - \\
    \hline
\end{tabular}
\end{table*}

\begin{table*}[t]
\caption{Error matrix on specific characters corresponding to the N-gram experiments}
\centering
\begin{tabular}{|c|c|c|c|c|c|c|}
	\hline
    Font & Character & Dataset Type & 2-gram & 3-gram & 4-gram & 5-gram \\
    \hline
    \multirow{4}{*}{Comic Bold} & \multirow{2}{*}{j} & Seen & 1.1\% & 0\% & 0\% & 0\% \\
    \cline{3-7}
    & & Purely Unseen & 6.5\% & 6.8\% & 5.6\% & 5.5\% \\
    \cline{2-7}
    & \multirow{2}{*}{w} & Seen & 7.8\% & 4.2\% & 3.0\% & 2.1\% \\
    \cline{3-7}
    & & Purely Unseen & 7.8\% & 7.4\% & 6.2\% & 6.8\% \\
    \hline
    \multirow{4}{*}{Arial Narrow} & \multirow{2}{*}{j} & Seen & 6.6\% & 3.5\% & 0\% & 0\% \\
    \cline{3-7}
    & & Purely Unseen & 29\% & 30.4\% & 29.4\% & 28.5\% \\
    \cline{2-7}
    & \multirow{2}{*}{t} & Seen & 1.9\% & 1.2\% & 1.1\% & 0.4\% \\
    \cline{3-7}
    & & Purely Unseen & 7.9\% & 5.0\% & 5.1\% & 5.4\% \\
    \hline
    \multirow{2}{*}{Times} & e & Seen & 70.5\% & 73.1\% & 72.2\% & 73.2\% \\
    \cline{2-7}
    & l & Seen & 20.6\% & 4.3\% & 1.2\% & 0.5\% \\
    \hline
\end{tabular}
\end{table*}

While the reasoning, of the above analysis is sound, it is not complete by itself. It is possible that fluctuations in the frequency of characters across test sets of different length may influence the experiments. To resolve any ambiguities from this, we inspect the performance on some characters across 2-gram to 5-gram datasets. The results are shown in Table 3.

We reassert results from the shuffled characters experiment by evaluating CER on \textit{Purely Unseen} 2-gram to 5-gram datasets. We omit 6 and 7-grams datasets for lack of sufficient samples. The N-grams in this case not only do not follow the language model seen in training, but also go out of their way in ensuring that any subsequence seen in training is not repeated while testing. Going with the complementary reasoning presented in the evaluation of \textit{Seen} experiments, we do not expect the performance to improve as length increases. The error should also stay consistently above the error on \textit{Seen} test sets. The results presented in Table 2 are consistent with our reasoning. We also perform and show results on \textit{Unseen} N-gram datasets, in Table 2. Consistent with previous reasoning it has consistently worse performance than \textit{Seen} datasets, but shows improvement with increasing N-gram length due to subsequences which have been \textit{Seen} in training.

\subsection{What about other fonts?}

The fonts highlighted in our experiments so far show improvement across all characters on \textit{Seen} sequences and therefore the overall performance measure is consistent with hypotheses across all \textit{Seen} N-gram test sets. However, that may not always be the case as we show with the third testing font Times Roman. The model has a proclivity for confusing only two characters in this font: \textit{l} gets confused for \textit{I} and \textit{e} for \textit{c} on the \textit{Seen} N-gram experiments. The performance for \textit{l} improves as N-increases dropping from 20.6\% to 0.5\% error, however the performance for \textit{e} stays approximately the same around 72\%. This exceptionally high error on a single character forces the results on any test set to be dictated by the frequency of \textit{e}. In order to highlight this point, we run another set of \textit{Seen} experiments where we recreate the datasets fixing the percentage of \textit{e} to be the same as that of 2-gram test set i.e. 6\%. We compare the results of both sets of experiments, where we regulate the final percentage composition of \textit{e} and where we do not in Table 4. The results become consistent for hypotheses once we regulate the rogue character \textit{e}. We inspect confusions on \textit{e} for a possible explanation on why it does not show any improvement, but we do not come across anything credible. The mistakes are spread across all font sizes and across different preceding and succeeding characters.

\begin{table}[t]
\caption{Differences in performance when frequency of \textit{e} is controlled in the times font}
\begin{center}
\begin{tabular}{|c|c|c|}
	\hline
    \multirow{3}{*}{N-gram} &  \multicolumn{2}{c|}{Times Roman} \\
    \cline{2-3}
    & \multicolumn{2}{c|}{Seen} \\
    \cline{2-3}
    & \textit{e} unregulated & \textit{e} regulated \\
    \hline
    2 & 5.6\% & 5.6\% \\
    \hline
    3 & 7.4\% & 4.7\% \\
    \hline
    4 & 9\% & 4.6\% \\
    \hline
    5 & 9.7\% & 4.6\% \\
    \hline
    6 & 10.1\% & 4.6\% \\
    \hline
    7 & 9.7\% & 4.6\% \\
    \hline
\end{tabular}
\end{center}
\end{table}

\section{Conclusion}

LSTM networks have been successful in OCR, but insight into what they learn for a given task is still lacking. We present evidence that LSTMs when trained for the OCR task, learn an implicit LM. We find that implicit LM improves performance up to 2.4\% CER when tested on synthetic English language data. As a real world problem extension it has also been shown that this implicit LM improves performance by up to 3.6\% CER on a multilingual OCR task \cite{ul-hasan_can_2013}. We also show that it makes use of up to 5 characters in making predictions. It does not necessarily help in making predictions on current character always as we saw with the indifference in performance on character \textit{e} in Times font. All experiments were conducted using English, but the general inference should hold good for any language.





%

\bibliographystyle{IEEEtran}
\bibliography{ImplicitLM}

\end{document}